\documentclass{article}

\usepackage{microtype}
\usepackage{graphicx}
\usepackage{subcaption}
\usepackage{booktabs} 

\usepackage{hyperref}

\usepackage[preprint]{icml2026}

\usepackage{amsmath}
\usepackage{amssymb}
\usepackage{mathtools}
\usepackage{amsthm}
\usepackage{multirow}

\usepackage{fancyvrb}
\usepackage{fvextra}

\DefineVerbatimEnvironment{myverbatim}{Verbatim}{%
	fontsize=\footnotesize,
	breaklines=true,
	breaksymbolleft={},
	frame=single,
}

\usepackage{xcolor}

\usepackage[capitalize,noabbrev]{cleveref}

\theoremstyle{plain}

\theoremstyle{definition}

\theoremstyle{remark}

\usepackage[T1]{fontenc}

\usepackage[textsize=tiny]{todonotes}

\icmltitlerunning{OpenGuanDan: A Large-Scale Imperfect Information Game Benchmark}

\begin{document}

\twocolumn[
  \icmltitle{OpenGuanDan: A Large-Scale Imperfect Information Game Benchmark}



  \icmlsetsymbol{equal}{*}

  \begin{icmlauthorlist}
    \icmlauthor{Chao Li}{equal,njupt}
    \icmlauthor{Shangdong Yang}{equal,njupt}
    \icmlauthor{Chiheng Zhan}{njupt}
    \icmlauthor{Zhenxing Ge}{nju}
    \icmlauthor{Yujing Hu}{netease}
    \icmlauthor{Bingkun Bao}{njupt}
    \\
    \icmlauthor{Xingguo Chen}{njupt}
    \icmlauthor{Yang Gao}{nju}
  \end{icmlauthorlist}

  \icmlaffiliation{njupt}{School of Computer Science, Nanjing University of Posts and Telecommunications}
  \icmlaffiliation{netease}{NetEase Fuxi AI Lab, Netease Inc}
  \icmlaffiliation{nju}{School of Intelligent Science and Technology, Nanjing University}

  \icmlcorrespondingauthor{Xingguo Chen}{chenxg@njupt.edu.cn}
  \icmlcorrespondingauthor{Yang Gao}{gaoy@nju.edu.cn}

  \icmlkeywords{Machine Learning, ICML}

  \vskip 0.3in
]



\printAffiliationsAndNotice{\icmlEqualContribution}

\begin{abstract}
  The advancement of data-driven artificial intelligence (AI), particularly machine learning, heavily depends on large-scale benchmarks. Despite remarkable progress across domains ranging from pattern recognition to intelligent decision-making in recent decades, exemplified by breakthroughs in board games, card games, and electronic sports games, there remains a pressing need for more challenging benchmarks to drive further research. To this end, this paper proposes OpenGuanDan, a novel benchmark that enables both efficient simulation of GuanDan (a popular four-player, multi-round Chinese card game) and comprehensive evaluation of both learning-based and rule-based GuanDan AI agents. OpenGuanDan poses a suite of nontrivial challenges, including imperfect information, large-scale information set and action spaces, a mixed learning objective involving cooperation and competition, long-horizon decision-making, variable action spaces, and dynamic team composition. These characteristics make it a demanding testbed for existing intelligent decision-making methods. Moreover, the independent API for each player allows human-AI interactions and supports integration with large language models. Empirically, we conduct two types of evaluations: (1) pairwise competitions among all GuanDan AI agents, and (2) human-AI matchups. Experimental results demonstrate that while current learning-based agents substantially outperform rule-based counterparts, they still fall short of achieving superhuman performance, underscoring the need for continued research in multi-agent intelligent decision-making domain. The project is publicly available at \url{https://github.com/GameAI-NJUPT/OpenGuanDan}.
\end{abstract}


\section{Introduction}
In recent years, data-driven artificial intelligence (AI), particularly machine learning methods, has witnessed remarkable advances across domains ranging from pattern recognition to intelligent decision-making. In the former, breakthroughs in computer vision (\emph{e.g.,} AlexNet, ResNet, YOLO, Segment Anything, and Depth Anything~\cite{krizhevsky2012imagenet,he2016deep,redmon2016you,kirillov2023segment,yang2024depth}) and natural language processing (\emph{e.g.,} BERT, GPT, Llama, DeepSeek~\cite{devlin2019bert,ouyang2022training,touvron2023llama,guo2025deepseek}) have brought unprecedented convenience to human life. These advances are largely driven by accessible large-scale data, such as vast collections of images~\cite{deng2009imagenet}, videos~\cite{caba2015activitynet}, and texts~\cite{raffel2020exploring}. 

A similar trend is also observed in the domain of intelligent decision-making, where agents are trained using massive amounts of interactive data generated within diverse benchmarks. Representative advancements include agents based on reinforcement learning (RL) and game theory methods, which have achieved, or even surpassed human-level performance in board games (\emph{e.g.,} Go, chess, and shogi~\cite{silver2016mastering,silver2017mastering,schrittwieser2020mastering}), card games (\emph{e.g.,} Texas Hold'em, mahjong, and DouDizhu~\cite{brown2018superhuman,brown2019superhuman,li2020suphx,zha2021douzero}), and electronic sports games (\emph{e.g.,} StarCraft II, Dota 2, and Honor of Kings~\cite{vinyals2019grandmaster,berner2019dota,ye2020towards}). In comparison to perfect information board games, where both agents\footnote{In this paper, we use agent and player interchangeably. The same convention applies to policy and strategy.} have full knowledge of the game state, imperfect information games, such as card and electronic sports games, restrict each agent's ability to perceive the complete state. For example, in card games, each agent cannot access other agents' private cards. Such imperfect information settings are prevalent in real-world scenarios including business, finance, and military applications, which underscores the need for novel game benchmarks that incorporate imperfect information and intricate complexity to advance research in multi-agent intelligent decision-making.

To this end, this paper proposes OpenGuanDan, a novel card game benchmark that incorporates two key components: (1) a highly efficient simulator of GuanDan, a popular Chinese card game involving two teams (of four players) competing across multiple rounds, and (2) a comprehensive evaluation of multiple GuanDan AI agents, including both learning-based and rule-based agents. OpenGuanDan constitutes a challenging benchmark due to the inherent complexity of GuanDan. Matches unfold over multiple rounds where two teams compete to exhaust their hand cards, with the finishing order determining level progression and dynamically influencing card strength in subsequent rounds. Such game mechanism gives raise to several nontrivial challenges, including imperfect information, large-scale information set and action spaces, a mixed learning objective involving cooperation and competition, long-horizon decision-making, variable action spaces, and dynamic team composition. For advancing data-driven AI research, in terms of simulation efficiency, OpenGuanDan supports extensive parallelism and achieves approximately 25 million steps per hour across 10 parallel environments on a single Intel Core i9-13900K CPU. Furthermore, in terms of evaluation, OpenGuanDan incorporates three learning-based agents (developed using RL and game theory methods) and four rule-based agents, alongside comparative results. Collectively, these attributes establish OpenGuanDan as a powerful testbed for advancing research in multi-agent intelligent decision-making.

To empirically assess the GuanDan AI agents within OpenGuanDan, we conduct two types of evaluations: (1) pairwise competitions among all agents, and (2) human-AI matchups. The results demonstrate that although current learning-based agents substantially outperform rule-based ones, they have not yet achieved superhuman performance. These findings highlight the considerable complexity posed by OpenGuanDan, and underscore the pressing need for further research in multi-agent intelligent decision-making domain.


\section{Related Work}
\label{sec:related_work}
In this section, we provide a concise review of existing game benchmarks, and AI agents developed using RL methods, game theory approaches, and large language model (LLM).

\begin{figure}[t]
	\centering
	\includegraphics[width=1.0\linewidth]{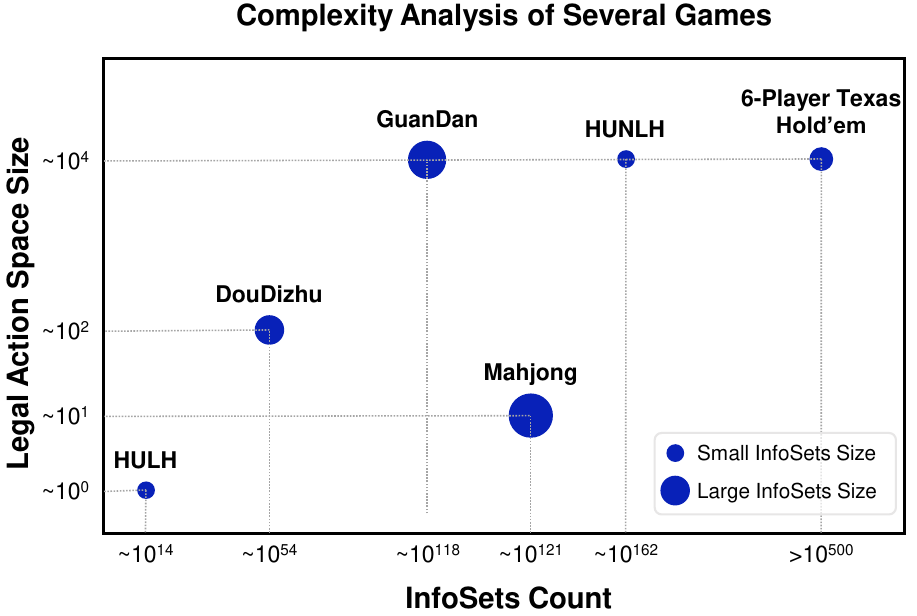}
	\caption{Game complexity of multiple representative card games. We report the information sets (infosets) count, the legal action space size, and the infosets size for Heads-Up Limit Texas Hold'em (HULH), Heads-Up No-Limit Texas Hold'em (HUNLH), Six-Player Texas Hold'em, DouDizhu, Mahjong, and GuanDan. Notably, the information sets size is approximately $10^{3}$ for HULH and HUNLH, $10^{15}$ for Six-Player Texas Hold’em, $10^{23}$ for DouDizhu, $10^{48}$ for Mahjong, and $10^{36}$ for GuanDan.}
	\label{fig:game_complexity_analysis}
\end{figure}

\textbf{Game Benchmarks.} Within the intelligent decision-making domain, games have long served as pivotal benchmarks for methodological development and empirical evaluation. For single-agent decision-making, representative games include Mountain Car~\cite{moore1990efficient}, Cart Pole~\cite{barto2012neuronlike}, Atari~\cite{bellemare2013arcade}, VizDoom~\cite{kempka2016vizdoom}, DeepMind Lab~\cite{beattie2016deepmind}, and Procgen~\cite{cobbe2020leveraging}, on which RL-based agents have demonstrated remarkable performance. In recent years, this success has extended from single-agent settings to multi-agent domains, leading to the emergence of a wide range of multi-agent game benchmarks. These include board games (\emph{e.g.,} Go, chess, and shogi~\cite{silver2016mastering,silver2017mastering,schrittwieser2020mastering}), card games (\emph{e.g.,} Texas Hold'em, Mahjong, Hanabi, and DouDizhu~\cite{brown2018superhuman,brown2019superhuman,li2020suphx,bard2020hanabi,zha2021douzero}), and electronic sports games (\emph{e.g.,} StarCraft II, Dota 2, Honor of Kings, Quake III Arena, SMAC, SMACv2, GRF, Fever Basketball, FightLadder ~\cite{vinyals2019grandmaster,berner2019dota,ye2020towards,jaderberg2019human,samvelyan2019starcraft,ellis2023smacv2,kurach2020google,jia2020fever,li2024fightladder}). Among them, card games, most notably poker, have traditionally served as the primary testbeds for solving imperfect information games.

Building on this line of research, this paper proposes OpenGuanDan, which introduces GuanDan as a novel and challenging poker game benchmark. As illustrated in Fig.~\ref{fig:game_complexity_analysis}, in comparison to existing card game benchmarks, GuanDan exhibits substantially increased difficulty, characterized by its high complexity across three critical dimensions: the count of information sets (up to $10^{118}$), the size of legal action space (up to $10^{4}$), and the size of information sets (up to $10^{36}$). These characteristics call for further methodological advances in solving imperfect-information games.

\textbf{Game AI Agents.} For multi-agent games, current AI agents can be divided into three major categories. The first category of agents leverages RL to learn effective policies. Representative successes include AlphaGo, AlphaZero, MuZero~\cite{silver2016mastering,silver2017mastering,schrittwieser2020mastering} for board games, Suphx~\cite{li2020suphx} for Mahjong, DouZero~\cite{zha2021douzero} for DouDizhu, AlphaStar~\cite{vinyals2019grandmaster} for StarCraft II, OpenAI Five~\cite{berner2019dota} for Dota 2, JueWu~\cite{ye2020towards} for Honor of Kings, FTW~\cite{jaderberg2019human} for Quake III Arena, and multi-agent RL approaches~\cite{lowe2017multi,sunehag2017value,rashid2020monotonic,yu2022surprising} for SMAC, SMACv2, GRF, and MPE. The second category primarily adopts game theory methods to compute efficient strategies. Representative examples include DeepStack~\cite{moravvcik2017deepstack} and Libratus~\cite{brown2018superhuman} for Heads-Up No-Limit Texas hold'em, as well as Pluribus~\cite{brown2019superhuman} for Six-Player No-Limit Texas Hold'em, all of which have exhibited superhuman performance. Although several RL-based agents for GuanDan, such as DanZero~\cite{lu2023danzero} and SDMC~\cite{ge2024solving}, as well as the GS2~\cite{ge2023efficient} agent based on game theory, have been proposed, we empirically demonstrate that these GuanDan AI agents fail to achieve superhuman performance in Sec.~\ref{sec:exp}.

The third category consists of agents developed based on LLM, which have demonstrated promising potential in card-game decision-making, such as Suspicion-Agent~\cite{guo2023suspicion}, PokerGPT~\cite{huang2024pokergpt}, Agent-Pro~\cite{zhang2024agent}, and ReTA~\cite{duan2024reta}. However, further performance improvements of LLM-based agents typically depend on fine-tuning with large-scale demonstration data, which can be generated more cost-effectively by RL-based and game-theoretic agents. As a result, continued research on RL and game theory methods remains essential for effectively addressing the challenges posed by GuanDan.

\section{GuanDan: A Chinese Card Game}
\label{sec:guandan}
In this section, we provide a detailed overview of the game mechanism of GuanDan as well as the associated challenges.

\subsection{Game Mechanism}
\textbf{Card Combinations.} GuanDan is played by four players organized into two teams, with teammates seated opposite each other. The game uses two standard decks comprising four suits, Spades, Hearts, Diamonds, and Clubs, with two copies of each ranking from $2$ to $K$ and $A$, along with four jokers (two red and two black). GuanDan supports a rich set of card combinations. (1) A \textit{single} consists of any individual card. (2) A \textit{pair} is formed by two cards of the same rank in any suits, except that a red joker and a black joker can not form a pair. (3) A \textit{triple} consists of three cards of the same rank. (4) A \textit{tube} comprises three consecutive pairs, and (5) a \textit{plate} consists of two consecutive triples. (6) A \textit{full house} is formed by combining a triple with an additional pair. (7) A \textit{straight} consists of five consecutive single cards, wherein the $A$ can act as either the highest or lowest rank, allowing sequences such as $A\kern0.05em\raisebox{0.1ex}{\scalebox{0.6}[1]{$\to$}}\kern0.05em 2\kern0.05em\raisebox{0.1ex}{\scalebox{0.6}[1]{$\to$}}\kern0.05em  3\kern0.05em\raisebox{0.1ex}{\scalebox{0.6}[1]{$\to$}}\kern0.05em 4\kern0.05em\raisebox{0.1ex}{\scalebox{0.6}[1]{$\to$}}\kern0.05em 5$ and $10\kern0.05em\raisebox{0.1ex}{\scalebox{0.6}[1]{$\to$}}\kern0.05em J\kern0.05em\raisebox{0.1ex}{\scalebox{0.6}[1]{$\to$}}\kern0.05em Q\kern0.05em\raisebox{0.1ex}{\scalebox{0.6}[1]{$\to$}}\kern0.05em K\kern0.05em\raisebox{0.1ex}{\scalebox{0.6}[1]{$\to$}}\kern0.05em A$. (8) Particularly powerful combinations include \textit{bomb}, defined as four or more cards of the same rank. (9) A \textit{straight flush} is a straight wherein all cards share the same suit. (10) The strongest is the \textit{joker bomb}, which consists of all four jokers.

\textbf{Level Cards and Wild Cards.} GuanDan features a multi-round level promotion mechanism, in which each round is associated with a specific level determined by the current promotion status of the leading team. Cards with the round level are referred as \textit{level cards} and are ranked above $A$ but below jokers. Among the level cards, the two heart-suited cards function as \textit{wild cards}. During a round, wild cards can substitute for any rank except jokers when forming combinations with other cards. However, when played individually (\emph{i.e.,} \textit{single}), they are treated as ordinary, non-wild cards. The detailed mechanism governing level progression across rounds is described in the subsequent game process.

\textbf{Card Combination Ranking.} For \textit{single}, the ranking order is red joker, black joker, \textit{level cards}, followed by $A$, and then ranks from $K$ down to $2$. \textit{Pairs} and \textit{triples} are ordered according to their card ranks. \textit{Tubes}, \textit{plates} and \textit{straights} are ordered based on the rank of their highest card. For
\textit{full house}, only the rank of the triple component is considered when determining superiority. All above card combinations are mutually incomparable, meaning that none can dominate another unless they belong to the same combination type.

\textit{Bomb} and \textit{straight flush} are strictly stronger than the aforementioned combinations and can be used to override them. Bombs are ranked primarily by their size, with bombs containing more cards always outranking those with fewer cards, regardless of rank. When bombs have the same size, they are further ordered by their ranks. \textit{Straight flushes} are treated as a special class of bombs: they outrank four- and five-card bombs but are inferior to bombs consisting of six or more cards, and are ranked based on the rank of their highest card, irrespective of suit. A graphical illustration of all card combinations and their rankings is provided in Appendix.~\ref{sec:card_combination_visualization}.

\textbf{Game Process.} A round of GuanDan begins with shuffling the deck and placing it face down. One player cuts the deck and reveals a card to determine the starting player, resulting in two cases. (1) If the revealed card is not a joker, the first drawer is selected by counting counterclockwise from the player who revealed the card. (2) Otherwise, the procedure is repeated. Gameplay then proceeds in a counterclockwise order. The lead player may play any valid card combination from their hand, while subsequent players must either play a higher-ranked card combination or choose to pass. If all other players pass, the most recent player to have played a combination gains the right to lead the next trick. When a player has emptied its cards and all other players pass, its teammate becomes the next leader; otherwise, the nearest player in counterclockwise order assumes the lead. At the end of each round, players are ranked according to the order in which they finish shedding their cards, yielding four roles: the \textit{Banker} (first), the \textit{Follower} (second), the \textit{Third} (third), and the \textit{Dweller} (fourth). Teams are defined by these roles; for example, a team whose members finish as the Banker and Follower is referred to as a \textit{Double-Winner} team.

\textbf{Level Progression.} In the multi-round level promotion, the level advancement of the winning team depends on the finishing positions of its members within a round. Specifically, three cases are distinguished: (1) If a team's players finish first (Banker) and second (Follower), the team advances by three levels. (2) If a team finishes first (Banker) and Third, the team advances by two levels. (3) If a team finishes first (Banker) and fourth (Dweller), the team advances by one level.
The losing team does not gain any levels in that round.

\textbf{Declaration, Tribute and Return.} In addition, during each round, players must declare when their hands contain ten or fewer cards. From the second round onward, a \textit{tribute} phase is introduced, during which the Dweller from the previous round must submit its highest-ranking card to the Banker, who then returns a card whose rank does not exceed $10$. In particular, when the winning team of the previous round is a \textit{Double-Winner} team (\emph{i.e.,} occupying both the Banker and Follower), both players of the losing team must pay tribute independently. Under this setting: (1) the highest-ranking of the two tribute cards is transferred to the Banker; (2) the second-highest tribute is transferred to the Follower; and (3) the return of cards is conducted within these corresponding player pairings. The player posing the highest-ranked tribute then assumes the lead. An exception applies when the losing team jointly holds both red jokers, in which case the tribute is waived and no card exchange occurs; the Banker assumes the lead. In the single-player tribute scenario, this exception applies when the single Dweller holds both red jokers. The tribute and return significantly affect card strength among players and introduce strong inter-round dependencies. In the following sections, we denote this exception as ``anti-tribute'' and the return process as ``back-tribute'' for clarity.

The match reaches its definitive conclusion when any team reaches Level $A$. To secure the final victory, the team must not only reach this level but also ensure that neither member occupies the Dweller in the final round. If a team at Level $A$ fails to avoid the Dweller three times in rounds played at its level, its level is reset to 2 and the progression restarts.


\subsection{Challenges}
The above game mechanism gives raise to several nontrivial challenges. (1) \textit{Imperfect information and large-scale information sets.} Players solely observe partial game states due to hidden hands and stochastic card distributions, resulting in expansive latent state spaces. Effective decision-making therefore requires belief modeling and opponent inference. (2) \textit{Large-scale and variable action spaces.} Legal actions are compositional and must be dynamically constructed as valid card combinations subject to combinatorial constraints imposed by players' cards, yielding an extremely large and state-dependent action space that varies across rounds. (3) \textit{Mixed cooperation-competition learning objective.} In GuanDan, agents must coordinate with a teammate while simultaneously competing against two opponents. This hybrid objective challenges learning frameworks designed for purely cooperative or purely adversarial settings. (4) \textit{Long-horizon decision-making.} In-round decisions affect finishing order, role assignments, and tribute mechanisms, inducing strong inter-round dependencies and long-horizon credit assignment challenges. (5) \textit{Dynamic team composition.} The level promotion mechanism and asymmetric round termination among teammates inherently induce dynamic team composition. Teammates may finish at different times; once one player empties their hand, the remaining teammate must act alone, effectively changing the team structure and required coordination. All of these challenges motivate the use of GuanDan as a challenging and well-suited testbed for research in multi-agent intelligent decision-making.

\section{Task Formalization}
\label{sec:task_formalization}
In this section, we formalize the GuanDan game from game theory and RL perspectives, respectively, and provide some recommendations on how research should be carried out.

\subsection{Game Theory Perspective}
\label{sec:game_theoretical_perspective}
\textbf{Formalization.} From the perspective of game theory, an imperfect information multi-agent task where agents take turns to make decision is formally modeled as an imperfect information extensive-form game $\langle N, H, \mathcal{I}, P, A, Z, \{\mu^{i}\}^{i\in N} \rangle$, where $N=\{1,2,\ldots,n\}\cup\{c\}$ denotes the player set and $c$ is the chance player. This formalization gives rise to a game tree, where $H$ is the set of all possible nodes (histories), represented as sequences of actions. For each node $h\in H$, the player function $P: H\rightarrow N$ specifies the acting player $P(h)$, and $A(h)$ defines the legal actions available to the player $P(h)$. Specially, the chance player $c$ chooses actions with a fixed probability. $Z\subseteq H$ is the set of terminal nodes. For each player $i$, the payoff function $u^{i}: Z\rightarrow \mathbb{R}$ determines the payoff that player $i$ will receive upon reaching a terminal node $z\in Z$. Imperfect information is represented by information sets (infosets), $\mathcal{I}$, which constitute a partition of all nodes $H$. For any infoset $I^{i}\in\mathcal{I}^{i}$, nodes $h, h'\in I^{i}$ are indistinguishable to the player $i$. Thus, the acting player and the corresponding legal actions must be identical across all nodes within an infoset. Formally, let $P(I^{i})$ and $A(I^{i})$ denote the acting player and legal actions in the infoset $I^{i}$, we have $P(I^{i})=P(h), A(I^{i})=A(h), \forall h\in I^{i}$.

A strategy $\sigma^{i}(I^{i})$ is a probability distribution over the legal actions $A(I^{i})$ for each infoset $I^{i}$ where $P(I^{i})=i$, and the probability of selecting the action $a$ is denoted as $\sigma^{i}(I^{i},a)$. A strategy profile $\sigma=(\sigma^{i}, \sigma^{-i})$ is a tuple of all players' strategies, including player $i$'s strategy $\sigma^{i}\in\sum^{i}$ and other players $-i$' strategies $\sigma^{-i}\in\sum^{-i}$. The expected payoff of player $i$ when all players follow the strategy profile $(\sigma^{i}, \sigma^{-i})$ is represented by $u^{i}(\sigma^{i}, \sigma^{-i})$. A Nash Equilibrium is a strategy profile $\sigma^{*}$ where no player can improve by unilaterally changing their strategies. That is, $\forall i\in N, u^{i}(\sigma^{*,i}, \sigma^{*,-i})\geq \max_{\sigma^{i}\in\sum^{i}}u^{i}(\sigma^{i}, \sigma^{*,-i})$. For each player $i$, a best response $BR(\sigma^{-i})$ is a strategy that maximizes its own payoff against other players' strategies $\sigma^{-i}$. Formally, $BR(\sigma^{-i})$ satisfies $BR(\sigma^{-i})=\max_{\sigma^{i}\in\sum^{i}}u^{i}(\sigma^{i}, \sigma^{-i})$.

\textbf{Possible Solutions.} For GuanDan, there are three situations regarding each player's strategy: (1) the Nash Equilibrium strategy, (2) the best response strategy against opponents' current strategies, and (3) the practical strategy with superior performance yet little theoretical guarantees. For case (1), due to the mixed learning objectives (\emph{i.e.,} cooperation and competition among four players) and decentralized strategies, the counterfactual regret minimization (CFR)~\cite{zinkevich2007regret} method and its variants, which specialize in two-player zero-sum games such as heads-up no-limit Texas hold'em, fail to guarantee the convergence to Nash Equilibrium strategies in multi-player games, yet still yielding superior performance~\cite{brown2019superhuman}. For case (2), although learning a best response to exploit opponents' weaknesses is appealing, accurately modeling opponent strategies in real time is often hard to achieve~\cite{schadd2007opponent,ganzfried2011game,yu2022model}. Consequently, we recommend to learn strategies following case (3), as done by Pluribus~\cite{brown2019superhuman}.

Specifically, we recommend subgame solving~\cite{burch2014solving,moravcik2016refining,brown2017safe} techniques for developing efficient strategies for GuanDan. Canonical subgame solving frameworks typically consist of two distinct processes: (1) offline computation of a coarse-grained blueprint strategy on an abstraction of the full game, and (2) online, real-time search in dynamically constructed subgames to further refine the blueprint strategy. Within the process (1), both information abstraction (where decision nodes with similar information are bucketed together and treated identically) and action abstraction (where the number of considered actions is reduced) are commonly used to reduce the game tree scale~\cite{sandholm2015abstraction,brown2015simultaneous}.
Subsequently, self-play methods, particularly CFR and its variants for poker, are applied to compute the blueprint strategy. Within the process (2), for large-scale imperfect information games (\emph{e.g.,} GuanDan), efficient real-time strategy refinement requires additional subgame reduction (further information abstraction over the information sets of both the current player and other players~\cite{ge2023efficient}), assumptions over multiple possible opponent strategies~\cite{brown2018superhuman}, and suitable equilibrium or approximation methods. Moreover, action-selection strategies must be carefully post-processed to maintain sufficient unpredictability, thus preventing exploitation by opponents.

\subsection{Reinforcement Learning Perspective}
\label{sec:rl_perspective}
\textbf{Formalization.} From the perspective of RL, the imperfect information issue can be regarded as a subset of the partial observability challenge. Accordingly, an imperfect information multi-agent task can be modeled as a partially observable stochastic game (POSG) $\langle N,S,\boldsymbol{A},\boldsymbol{R},P,\gamma,\boldsymbol{Z},\boldsymbol{O} \rangle$, where $N=\{1,2,\ldots,n\}$ denotes the agent set and $S$ is the state space. $\boldsymbol{A}=A^{1}\times A^{2} \times \ldots A^{n}$ represents agents' joint action space and $A^{i}$ denotes the local action space of agent $i\in N$. At each time step $t$, each agent $i$ receives its local observation $o_{t}^{i}\in Z^{i}$, drawn according to its observation function $O^{i}(o_{t}^{i}|s_{t})$, where $s_{t}$ represents the state and $\boldsymbol{Z}=\{Z^{i}\}^{i\in N},\boldsymbol{O}=\{O^{i}\}^{i\in N}$. Then, each agent $i$ selects its local action $a_{t}^{i}$ according to its policy $\pi^{i}$. Given the joint action of all agents $\boldsymbol{a}_{t}=(a_{t}^{1}, a_{t}^{2}, \ldots, a_{t}^{n})$, the environment transits to the next state $s_{t+1}$ following the state transition function $P(s_{t+1}|s_{t},\boldsymbol{a}_{t})$, and each agent $i$ receives its local reward $r_{t}^{i}$ specified by its reward function $R^{i}(s_{t},\boldsymbol{a}_{t})\in \boldsymbol{R}$. To deal with the partial observability, each agent $i$ conditions its policy $\pi^{i}(a_{t}^{i}|\tau_{t}^{i})$ on its local action-observation history $\tau_{t}^{i}=(o_{0}^{i},a_{0}^{i},o_{1}^{i},\ldots,o_{t}^{i})$. The objective of each agent $i$ is to learn the optimal policy $\pi^{i,*}$ that maximizes the expected discounted return $\mathbb{E}_{\boldsymbol{\pi},P}[\sum_{t=0}^{\infty}\gamma^{t}r_{t}^{i}]$, where $\gamma$ is the discount factor and $\boldsymbol{\pi}=(\pi^{1},\pi^{2},\ldots,\pi^{n})$ is agents' joint policy.

In comparison to classical POSG, where all agents simultaneously select actions and both the transition and reward functions are conditioned on agents' joint action, GuanDan allows only a single agent to act at each time step. To accommodate this game mechanism within the POSG framework, we define the joint actions $\boldsymbol{a}_{t}$, such that agents that do not act at time $t$, or that have already exhausted their hand cards, are assigned a NULL action (\emph{i.e.,} a zero vector). This formulation ensures compatibility between the standard POSG formalization and the turn-based dynamics of GuanDan.

\textbf{Possible Solutions.} For multi-agent games, empirical evidence suggests that RL-based policies can achieve strong performance in practice, albeit with limited theoretical guarantees. Therefore, at present, we recommend learning RL-based policies under case (3) described from the game theory perspective, while leaving the investigation of formal convergence and optimality guarantees for future work. In terms of RL-based policy learning, we advocate the Deep Monte Carlo (DMC) method, which has shown efficiency in dealing with large-scale imperfect information card game such as DouDizhu, as exemplified by DouZero~\cite{zha2021douzero}. In general, DMC-based agents for card games consist of three key components: (1) efficient representations of each agent's observations and the set of legal actions at each time step, (2) an efficient self-play training paradigm to generate high-quality interactive data, and (3) decision policies with low exploitability during action selection.

For component (1), each agent should encode information including, but not limited to, its private cards, the number of wild cards, the cards historically played by all agents, the remaining card counts of all agents, and the current level information. Such representations are crucial for mitigating the partial observability challenge. For component (2), naive self-play often produces low-quality or uninformative data during the early training process, when agents behave nearly randomly. Therefore, additional mechanisms are required to accelerate the transition from random to competent policies, such as incorporating teacher or expert policies to guide early self-play~\cite{ge2024solving}. Finally, for component (3), maintaining an appropriate level of stochasticity during action selection can reduce policy predictability and thereby mitigate the risk of exploitation by opponents.

\section{OpenGuanDan}
\label{sec:openguandan}
In this section, we provide a detailed explanation of OpenGuanDan. We first introduce our self-developed GuanDan simulator, in which predefined, and optionally customizable, wrappers for observation, action, and reward are formulated from a RL perspective. Subsequently, we present the design of independent per-player APIs and describe the technical implementation underlying the highly efficient simulation. Finally, we incorporate several GuanDan AI agents, including both learning-based and rule-based agents, and briefly introduce them for subsequent empirical comparisons.

\subsection{GuanDan Simulator}
The self-developed simulator faithfully implements the complete GuanDan game process under standard rules, including card dealing, tribute, anti-tribute, back-tribute, card playing, and level progression. During each round, the simulator broadcasts to every player, in a JSON-based format, comprehensive game information such as the player's hand cards, the set of legal actions, the cards played by other players, the levels, and termination signals. An entire GuanDan match consists of multiple rounds and terminates when one team reaches level $A$, with its two members assuming the roles of Banker and Follower. For clarity, a code-level demonstration of a complete GuanDan game is provided in Appendix~\ref{sec:guandan_simulator_appendix}.

\textbf{Observation.} Based on the information broadcasted by the simulator, we design an observation wrapper for each player within OpenGuanDan. Specifically, for each player $i$, the definition of its observation contains information as follows:

(1) player $i$'s private hand cards;

(2) the remaining unseen cards, excluding player $i$'s hand cards and all cards that have been played;

(3) all cards played by the Left-Hand Opponent (LHO);

(4) all cards played by the partner;

(5) all cards played by the Right-Hand Opponent (RHO);

(6) the action taken by the immediately preceding player;

(7) the most recent action taken by the LHO;

(8) the most recent action executed by the partner;

(9) the most recent action taken by the RHO;

(10) the number of remaining cards held by the LHO;

(11) the number of remaining cards held by the partner;

(12) the number of remaining cards held by the RHO;

(13) the levels of player $i$'s team and the opposing team, as well as the level of the current round;

(14) the number of wild cards in player $i$'s private cards.

GuanDan is played with two standard decks of cards (\emph{i.e.,} each deck comprises 4$\times$13+2=54 cards). Accordingly, for terms (1), (2), (3), (4), and (5), we employ a 54-dimensional vector for each of them, where each entry takes values in $\{0, 1, 2\}$ to indicate the count of each card type. For terms (10), (11), and (12), we assign them 28-dimensional one-hot vectors to represent the remaining card counts of the other three players, respectively, which range from 0 to 27. Terms (6), (7), (8), and (9) are encoded using 79-dimensional vectors that capture the card composition, the type of card combination, and the precedence order among combinations. Given that each team can progress through 13 possible levels, term (13) is denoted by a 39-dimensional vector. Finally, term (14) is encoded as a 13-dimensional vector, with each element in $\{0,1,2\}$ indicating the number of wild cards associated with the current round.

At each time step, the GuanDan simulator provides all available information to each player. This allows researchers to flexibly define their own customized observations by implementing wrappers that extract certain information.

\textbf{Action.} Considering that GuanDan involves three distinct decision-making stages (\emph{i.e.,} tribute, back-tribute, and card play), we design three corresponding sub-modules for generating per-agent legal actions, which are detailed as follows:

(1) \emph{Tribute Sub-Module.} For a player required to pay tribute, OpenGuanDan provides a tribute sub-module that follows the standard game rules by selecting the card with the highest rank as the tribute card. However, when multiple cards possess the same rank but differ in suit within the player's private cards, the choice of suit becomes nontrivial, as certain suits may enhance the card strength of the receiving player. To address this issue, researchers may either learn a dedicated tribute policy or design more sophisticated rule-based strategies for the selection of tribute cards.

(2) \emph{Back-Tribute Sub-Module.} For a player performing back-tribute, OpenGuanDan includes a back-tribute sub-module that returns all legally available back-tribute cards. However, the rank and suit of the back-tribute card may significantly influence the strength of the receiving player's hand cards. To avoid this, effective back-tribute policies or carefully designed heuristic rules are required for back-tribute players.

(3) \emph{Card Play Sub-Module.} For each player, the card play sub-module enumerates all legal card combinations as candidate actions, according to the player's private cards and the cards (actions) played by the immediately preceding player. Such explicit identification of legal actions typically facilitates efficient policy learning for GuanDan AI agents.

\textbf{Reward.} At the end of each round, players in the winning team receive rewards of +3, +2, and +1 when the Banker's partner is the Follower, the Third, and the Dweller, respectively, while players in the losing team are assigned rewards of -3, -2, and -1. When the level of the final round of a game reaches $A$, the reward is set to 0 if the Banker's partner is the Dweller, since the Banker's team cannot win the game under this condition. This reward design follows the setting adopted in DanZero~\cite{lu2023danzero}, which empirically demonstrates efficiency in facilitating policy learning.

\textbf{API.} The GuanDan simulator provides independent action-upload API for each player, facilitating efficient human-AI evaluation and seamless integration of diverse GuanDan AI agents developed based on game theory, RL, and LLM, as well as supporting self-play training. Example code snippets demonstrating action submission during the card-play, tribute, and back-tribute phases are provided in Appendix.~\ref{sec:api_design}.

What's more, the GuanDan simulator provides built-in game visualization, with a representative game snapshot shown in Fig.~\ref{fig:GuanDan_simulator_snapshot} (Appendix.~\ref{sec:guandan_simulator_appendix}). This functionality allows researchers to directly inspect and analyze the policies of diverse GuanDan AI agents by loading them through the player API. In addition, game players can be controlled using the mouse, enabling convenient and intuitive human-AI evaluation.

\begin{figure}[t]
	\centering
	\includegraphics[width=0.85\linewidth]{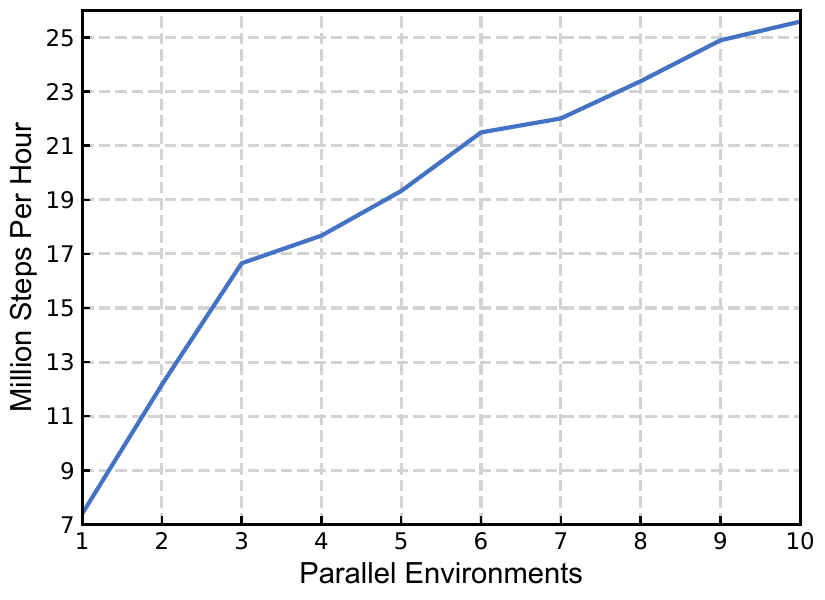}
	\caption{Number of simulation steps per hour versus number of parallel environments for the GuanDan simulator on a machine with 24-core Intel Core i9-13900K CPU (3.00GHz).}
	\label{fig:simulation_per_hour}
\end{figure}

\textbf{Technical Implementation.} The self-developed GuanDan simulator is implemented in highly optimized Java code, enabling highly efficient simulation on standard CPU-based machines. This allows a throughput of approximately 25 million steps per hour across 10 parallel environments on a 24-core Intel Core i9-13900K CPU (3.00GHz) (see Fig.~\ref{fig:simulation_per_hour}).

\subsection{Built-In Agents}
In addition to the GuanDan simulator, OpenGuanDan provides a collection of built-in AI agents to facilitate fair and reproducible comparisons when developing new GuanDan AI agents. This repository includes both learning-based and rule-based agents. Below, we briefly introduce each agent.

\textbf{GS2.} Following the subgame solving technique in Sec.~\ref{sec:game_theoretical_perspective}, GS2~\cite{ge2023efficient} further reduces the subgame by two processes: (1) reducing the number of the traversing player's infosets $|\mathcal{I}^{i}|$ based on the knowledge-limited subgame solving method~\cite{zhang2021subgame}, and (2) reducing the number of other players $-i$' infosets $|\mathcal{I}^{-i}|$ by learning a generation function to identify a subset of other players' infosets. In addition, a diversity-based generation function is introduced to theoretically and empirically reduce the exploitability of the refined strategy derived from the reduced subgame. Collectively, these techniques make GS2 applicable to large-scale imperfect information games with strong performance yet significantly reduced computational cost.

\textbf{DanZero.} In contrast, DanZero~\cite{lu2023danzero} belongs to RL-based GuanDan AI agents. Specifically, DanZero first designs specialized state and action encoding schemes, and then applies a deep Monte Carlo method combined with a self-play training paradigm to approximate the Q values of state-action pairs.To accelerate learning, DanZero adopts distributed training across multiple parallel GuanDan games. Empirically, DanZero achieves competitive performance on a normal server after approximately 30 days of training.

\textbf{SDMC.} Similar to DanZero, SDMC~\cite{ge2024solving} employs a deep Monte Carlo method together with self-play to estimate Q values for all state-action pairs, using customized state and action representations. Distinctively, SDMC accelerates the early stages of self-play training by incorporating an expert-level policy to guide per-player action selection. In addition, SDMC employs soft action sampling, selecting actions whose Q values are close to the maximum rather than always choosing the single best action, which effectively reduces the exploitability of the learned policy.

\begin{table*}[t]
	\caption{The pair-wise comparison results among all GuanDan AI agents.}
	\label{tab:pair_wise_comparison_results}
	\centering
	\resizebox{1.0\linewidth}{!}{
		\begin{tabular}{c|c|c|c|c|c}
			\toprule
			Agent & Opponent Agent & Percentage of Rounds with 3 Rewards (\%) & Percentage of Rounds with 2 Rewards (\%) & Percentage of Rounds with 1 Reward (\%) & Win Rate (\%) \\
			\midrule
			SDMC & \multirow{6}{*}{GS2} & 22.4 & 10.2 & 14.4 & 43.3 \\
			
			DanZero	&  & 20.7 & 7.8 & 20.8 & 38.0 \\
			
			Rule One &  & 9.3 & 9.6 & 12.2 & 7.0 \\
			
			Rule Two &  & 7.7 & 8.2 & 14.3 & 2.9 \\
			
			Rule Three &  & 7.3 & 8.9 & 14.3 & 2.0 \\
			
			Rule Four &  & 3.8 & 5.6 & 10.3 & 0.0 \\
			
			\midrule
			GS2	& \multirow{6}{*}{DanZero} & 26.6 & 13.6 & 10.5 & 62.0 \\
			
			SDMC &  & 27.0 & 12.4 & 10.6 & 57.6 \\
			
			Rule One &  & 12.2 & 13 & 11.7 & 16.8 \\
			
			Rule Two &  & 10.9 & 11.8 & 13.2 & 14.6 \\
			
			Rule Three &  & 9.6 & 11.4 & 12.0 & 9.8 \\
			
			Rule Four &  & 7.9 & 10.6 & 11.4 & 7.7 \\
			
			\midrule
			GS2	& \multirow{6}{*}{SDMC} & 24.7 & 11.5 & 16.8 & 56.7 \\
			
			DanZero	&  & 22.0 & 7.1 & 20.9 & 42.4 \\
			
			Rule One &  & 10.9 & 10.5 & 12.4 & 8.8 \\
			
			Rule Two &  & 9.9 & 12.1 & 13.1 & 4.2 \\
			
			Rule Three &  & 8.3 & 12.6 & 13 & 3.6 \\
			
			Rule Four &  & 5.0 & 6.3 & 10.2 & 0.0 \\
			
			\midrule
			GS2	& \multirow{6}{*}{Rule One} & 42.8 & 15.3 & 10.8 & 93.0 \\
			
			SDMC &  & 43.5 & 12.8 & 9.9 & 91.2 \\
			
			DanZero	&  & 38.2 & 14.3 & 10.6 & 83.2 \\
			
			Rule Two &  & 25.3 & 9.2 & 13.8 & 49.6 \\
			
			Rule Three &  & 9.1 & 14.8 & 24.1 & 42.2 \\
			
			Rule Four &  & 16.7 & 5.7 & 23.0 & 32.7 \\
			
			\midrule
			GS2	& \multirow{6}{*}{Rule Two} & 46.4 & 14.2 & 9.2 & 97.1 \\
			
			SDMC &  & 46.4 & 10.3 & 8.2 & 95.8 \\
			
			DanZero	&  & 40.4 & 12.3 & 11.4 & 85.4 \\
			
			Rule One &  & 29.3 & 5.8 & 16.5 & 50.4 \\
			
			Rule Three &  & 17.3 & 9.0 & 13.7 & 41.6 \\
			
			Rule Four &  & 16.6 & 14.3 & 15.0 & 18.8 \\
			
			\midrule
			GS2	& \multirow{6}{*}{Rule Three} & 47.7 & 13.2 & 8.6 & 98.0 \\
			
			SDMC &  & 49.2 & 10.3 & 6.6 & 96.4 \\
			
			DanZero	&  & 45.1 & 11.3 & 10.6 & 90.2 \\
			
			Rule One &  & 28.7 & 11.4 & 18.4 & 57.8 \\
			
			Rule Two &  & 25.0 & 7.0 & 28.1 & 58.4 \\
			
			Rule Four &  & 22.9 & 9.2 & 21.8 & 46.3 \\
			
			\midrule
			GS2	& \multirow{6}{*}{Rule Four} & 68.7 & 7.4 & 4.2 & 100.0 \\
			
			SDMC &  & 63.7 & 10.6 & 4.2 & 100.0 \\
			
			DanZero	&  & 51.9 & 8.1 & 10.1 & 92.3 \\
			
			Rule One &  & 29.2 & 10.5 & 14.4 & 67.3 \\
			
			Rule Two &  & 37.1 & 8.1 & 9.5 & 81.2 \\
			
			Rule Three &  & 29.1 & 5.3 & 8.3 & 53.7 \\
			\bottomrule
		\end{tabular}
	}
\end{table*}

In addition, OpenGuanDan includes four rule-based agents released from a Chinese GuanDan AI competition, referred to as Rule One through Rule Four. All of them are developed based on handcrafted domain knowledge of GuanDan. We include them to validate the ability of learning-based agents to acquire effective strategies solely from interactive data, and to serve as lightweight baselines for the preliminary evaluation of newly developed GuanDan AI agents.

\section{Experiments}
\label{sec:exp}
In this section, we design experiments to answer the following three questions: (1) Which GuanDan AI agents exhibit the strongest overall performance against all other agents? (See Sec.~\ref{sec:pair_wise_evaluation}) (2) Which levels of human GuanDan expertise are comparable to the performance of these agents? (See Sec.~\ref{sec:human_ai_evaluation}) (3) What is the computational time cost incurred by each agent during action selection? (See Appendix.~\ref{sec:inference_time_evaluation})

\begin{table*}[t]
	\caption{The human-AI evaluation results of all learning-based GuanDan AI agents. Due to the volunteers' limited familiarity with the simulator, the reported win rates may be slightly overestimated because of operational errors during evaluation.}
	\label{tab:human_ai_evaluation_results}
	\centering
	\resizebox{1.0\linewidth}{!}{
		\begin{tabular}{c|c|c|c|c|c}
			\toprule
			Agent & Total Rounds & Overall Win Rate (\%) & Win Rate against Beginners (\%) & Win Rate against Intermediate Players (\%) & Win Rate against Advanced Players (\%) \\
			\midrule
			GS2 & 106 & 42.5 (45/106) & 53.8 (7/13) & 40 (24/60) & 42.4 (14/33) \\
			
			SDMC & 106 & 39.6 (42/106) & 66.7 (8/12) & 37.9 (25/66) & 32.1 (9/28) \\
			
			DanZero	& 102 & 34.3 (35/102) & 43.2 (16/37) & 36.4 (12/33) & 21.9 (7/32) \\
			\bottomrule
		\end{tabular}
	}
\end{table*}

\subsection{Pair-Wise Evaluation}
\label{sec:pair_wise_evaluation}
For question (1), we conduct pair-wise comparisons among all GuanDan AI agents, that is, the players in the same team are controlled by a single GuanDan AI agent. For each agent pair, we conduct 1000 games and report the percentage of rounds with 3 rewards, 2 rewards, 1 reward, and the overall win rate. The comparison results are illustrated in Tab.~\ref{tab:pair_wise_comparison_results}. 

It is obvious that, the learning-based agents, including GS2, SDMC, and DanZero, demonstrate clear dominance over all rule-based agents. Specifically, their win rates exceed 80\% against Rule One and Two, and 90\% against Rule Three and Four, wherein rounds with 3 rewards constitute the majority of their victories. This performance gap is attributed to the lack of multiple critical decision-making scenarios within the design of these rule-based agents. In general, manually designing a rule-based GuanDan AI agent that is capable of enumerating and handling all possible decision situations is challenging and prohibitively expensive due to the complex game mechanism of GuanDan. In contrast, given the large volume of interactive data generated within the GuanDan simulator, data-driven AI techniques, including the subgame solving and DMC methods mentioned in Sec.~\ref{sec:task_formalization}, are able to learn effective strategies from self-play data. These results further demonstrate the effectiveness of subgame solving and DMC methods in addressing the challenges posed by large-scale imperfect information card games.

Among the learning-based agents, both SDMC and DanZero are developed based on the DMC method and the self-play training paradigm. What's different is that SDMC employs expert-level strategies to accelerate the early-stage self-play training and introduces soft action sampling to reduce exploitability during action selection. Consequently, SDMC outperforms DanZero, achieving a 57.6\% win rate against DanZero and better performance against other agents.

Furthermore, GS2 adopts the SDMC strategy as a blueprint and refines it through real-time subgame solving, effectively combining learned strategies with online search. As a result, GS2 significantly outperforms both SDMC and DanZero, achieving a 56.7\% win rate against SDMC and a 62\% win rate against DanZero. This further highlights the effectiveness of augmenting learned strategies with real-time search in solving large-scale imperfect information game.

\subsection{Human-AI Evaluation}
\label{sec:human_ai_evaluation}
To further evaluate the performance of learning-based GuanDan AI agents against human players, we host a human-AI competition involving 16 volunteers organized into 8 teams according to their GuanDan expertise. Specifically, 2 teams consist of beginners with less than 1 year of experience, 3 teams comprise intermediate players with 1-2 years of experience, and the remaining teams include advanced players with more than 3 years of experience. Each team independently selects any AI agent as opponent, and we ensure that each AI agent participates in approximately 100 rounds in total. For higher-performing agents (GS2 and SDMC), we deliberately schedule more matches against intermediate and advanced players and fewer matches against beginners.

The results are summarized in Tab.~\ref{tab:human_ai_evaluation_results}. None of the evaluated agents achieves an overall win rate above 50\%, indicating that they have not yet reached superhuman performance. In particular, GS2 and SDMC achieve win rates above 50\% against beginners, but below 45\% against intermediate and advanced players, which suggests that their playing strength lies between beginner and intermediate levels. DanZero achieves a 43.2\% win rate against beginners, 36.4\% against intermediate players, and only 21.9\% against advanced players, indicating that it fails to consistently outperform even beginner-level human players. Overall, these results demonstrate that current learning-based GuanDan AI agents remain below human expert-level performance, underscoring the need for continued research and algorithmic advancements.

\section{Conclusion}
\label{sec:conclusion}
This paper presents OpenGuanDan, a large-scale imperfect information card game benchmark that enables both efficient simulation of GuanDan and comprehensive evaluation of multiple GuanDan AI agents. Empirical results demonstrate that, although learning-based agents outperform rule-base agents, they remain far from superhuman performance. This highlights the substantial challenges posed by OpenGuanDan and underscores the need for continued research.

\textbf{Limitation and Future Work.} We identify three limitations that warrant further exploration. First, the tribute and back-tribute sub-modules necessitate more effective policies to further enhance agent performance. Second, to improve the simulation efficiency of GuanDan, we plan to optimize the code implementations of OpenGuanDan. Third, the current human-AI evaluation lacks participation from professional players. We plan to host a competition with prize incentives to attract expert players. We leave them as our future work.



\bibliography{reference}
\bibliographystyle{icml2026}

\newpage
\appendix
\onecolumn
\section{Card Combination Visualization}
\label{sec:card_combination_visualization}
\begin{figure}[H]
	\centering
	\includegraphics[width=1.0\linewidth]{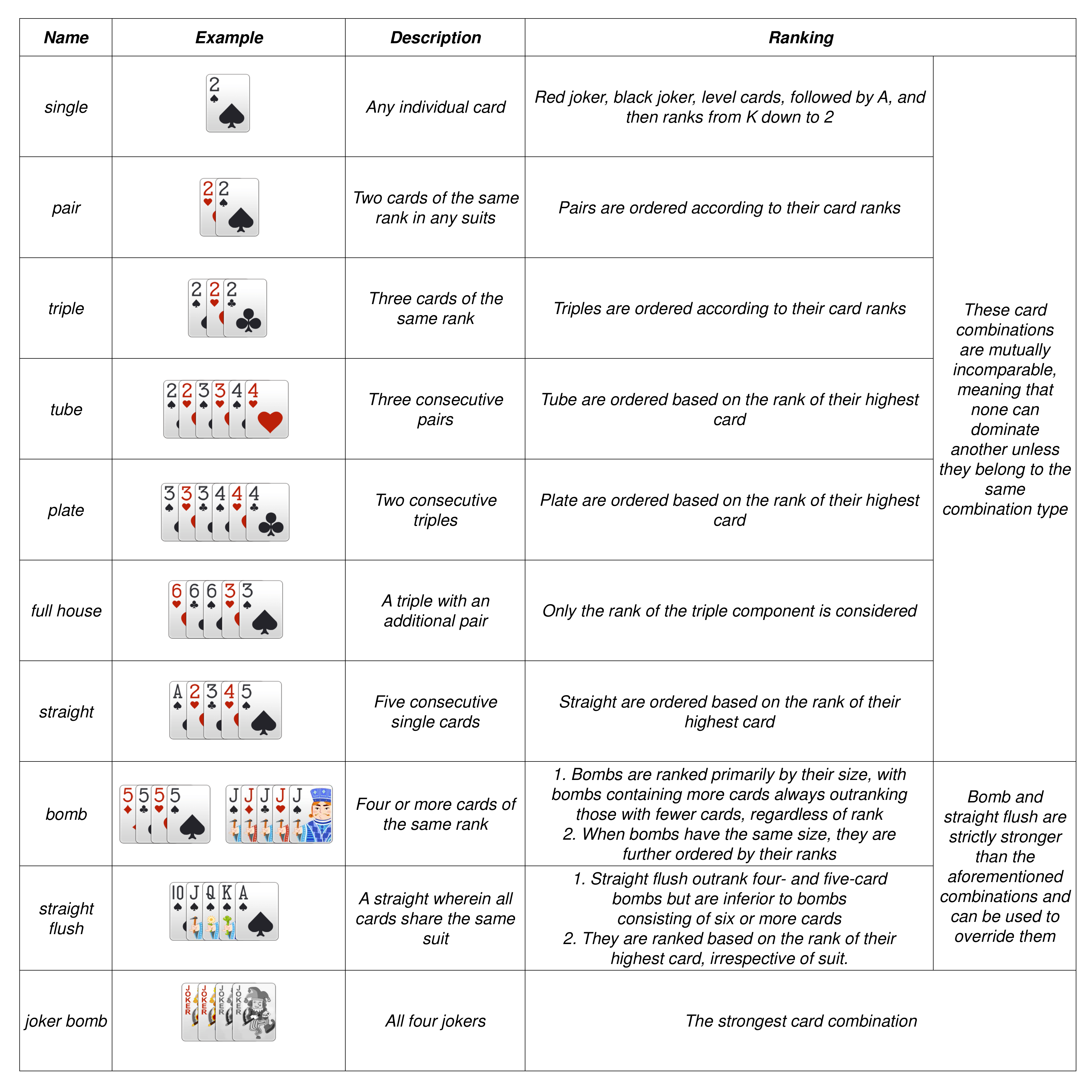}
	\caption{Visualization of all card combinations and their rankings.}
	\label{fig:card_combination}
\end{figure}

A graphical illustration of all card combinations, including \textit{single}, \textit{pair}, \textit{triple}, \textit{tube}, \textit{plate}, \textit{straight}, \textit{full house}, \textit{bomb}, \textit{straight flush}, and \textit{joker bomb}, is depicted in Fig.~\ref{fig:card_combination}. Regarding their rankings, note that the first 7 card combinations (\textit{single}, \textit{pair}, \textit{triple}, \textit{tube}, \textit{plate}, \textit{straight}, \textit{full house}) are mutually incomparable, meaning that none can dominate another unless they belong to the same combination type. \textit{Bomb} and \textit{straight flush} are strictly stronger than the aforementioned combinations and can be used to override them. Specially, \textit{straight flush} is treated as a special class of bombs: they outrank four- and five-card bombs but are inferior to bombs consisting of six or more cards. \textit{Joker bomb} is the strongest card combination.

\section{GuanDan Simulator}
\label{sec:guandan_simulator_appendix}
\begin{figure}[h]
	\centering
	\includegraphics[width=0.8\linewidth]{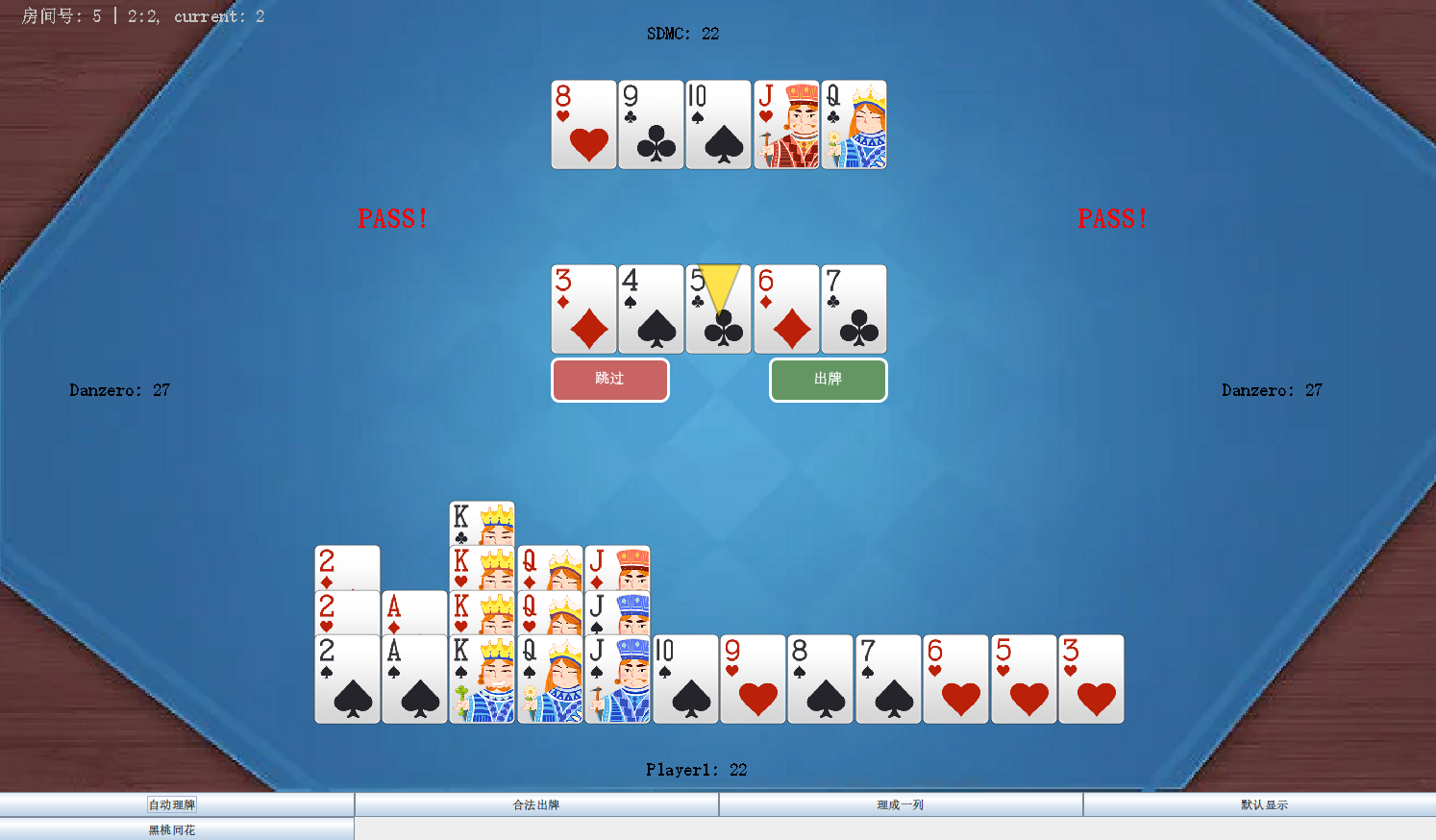}
	\caption{A snapshot of the self-developed GuanDan simulator, which supports both Chinese and English.}
	\label{fig:GuanDan_simulator_snapshot}
\end{figure}

In this section, we provide a step-by-step introduction to the code-level implementation of a complete GuanDan game within our self-developed GuanDan simulator, as depicted in Fig.~\ref{fig:GuanDan_simulator_snapshot}. We begin by describing the messages transmitted from the client to the server, including \textit{create room}, \textit{join room}, \textit{play action}, \textit{tribute action}, and \textit{back-tribute action}. Subsequently, we introduce the messages sent from the server to the client, including \textit{game start}, \textit{play notification}, \textit{tribute notification}, \textit{anti-tribute notification}, \textit{back-tribute notification}, \textit{episode over}, \textit{game over}, and \textit{game result}. Finally, we present the detailed message formats for \textit{card play}, \textit{tribute}, and \textit{back-tribute}.

\textbf{Create Room.} A complete GuanDan game begins with the creation of a room, where the player ID, the maximum number of rounds, and the seat index of the current player must be specified. The corresponding message format is shown below:
\begin{myverbatim}
'type': 'CREATE_ROOM',
'data': {
	'userId': 'user1', # User ID
	'round': 1,	# Number of rounds
	'seatNum': 0	# Seat number
}
\end{myverbatim}

\textbf{Join Room.} After a game room is created, a player can join the game by specifying the player ID, the room ID, and the target seat index. Once the number of connected players reaches 4, the game is automatically initialized and started by the server. The message format for joining a room is shown below:
\begin{myverbatim}
type': 'JOIN_ROOM',
'data': {
	'userId': 'user2',	# User ID
	'roomId': 1,	# Room ID
	'seatNum': 1	# Seat number
}
\end{myverbatim}

\textbf{Play Action.} During a GuanDan game, each player submits its selected action (\emph{i.e.,} card combination) to the server. Here, \texttt{act} denotes an action tuple and must be chosen from the set of legal actions provided by the server at the current decision step. The corresponding message format is as follows:
\begin{myverbatim}
'type': 'PLAY',
'data': {
	'roomId': 1,	# Room ID
	'player': 0,	# Player seat number
	'act': ['PASS', 'PASS', 'PASS']		# Action tuple
}
\end{myverbatim}

\textbf{Tribute Action.} Similar to the play action, during the tribute phase, the designated tribute player uploads its selected card(s) to the server. The tribute action \texttt{act} must be selected from the set of legal actions provided by the server. The corresponding message format is shown below:
\begin{myverbatim}
'type': 'TRIBUTE',
'data': {
	'roomId': 1,	# RoomID
	'player': 0,	# Player seat number
	'act': ['tribute', 'tribute', ['D2']]	# Tribute action
}
\end{myverbatim}

\textbf{Back-Tribute Action.} After receiving the tribute card (\texttt{tribute}) from the player seated at \texttt{tributePos}, the back-tribute player specifies its selected card(s) in \texttt{act} and submits them to the server. The back-tribute action must also be chosen from the set of legal actions provided by the server. The corresponding message format is as follows:
\begin{myverbatim}
'type': 'PAYTRIBUTE',
'data': {
	'roomId': 1,	# RoomID
	'player': 0,	# Player seat number
	'tributePos': 3,	# Seat number of the player who paid tribute
	'tribute': 'S2',	# The card offered as tribute
	'act': ['back', 'back', ['H2']]		# Back-tribute action
}
\end{myverbatim}

Below, we present the notification messages broadcast by the server to all connected players.

\textbf{Game Start.} At the beginning of each game, each player (identified by the seat index \texttt{myPos}) receives its private hand information (\texttt{handCards}) from the server. The corresponding notification message is shown as follows:
\begin{myverbatim}
'type': 'notify',
'stage': 'beginning',
'handCards': ['S2', 'H2', 'C2', ...],	# Hand cards list
'myPos': 1		# Player seat number
\end{myverbatim}

\textbf{Play Notification.} During the play phase, when the current player at seat \texttt{curPos} executes an action, the server broadcasts the current action and the highest-ranked action in the ongoing trick. Specifically, \texttt{greaterPos} denotes the seat index of the player who currently holds the highest action, and \texttt{greaterAction} records the corresponding card combination. The message format is given below:
\begin{myverbatim}
'type': 'notify',
'stage': 'play',
'curPos': 1,	# Current player's seat number
'curAction': ['Single', '2', ['S2']],	# Current player's action
'greaterPos': 1,	# Seat number of player with the highest action
'greaterAction': ['Single', '2', ['S2']]	# Highest action
\end{myverbatim}

\textbf{Tribute Notification.} After the tribute phase is completed, the server broadcasts the tribute outcomes to all players. The field \texttt{result} records the tribute transactions between players, where each entry specifies the tribute-paying player, the receiving player, and the corresponding card.
\begin{myverbatim}
'type': 'notify',
'stage': 'tribute',
'result': [[0, 3, 'S2']]	# Tribute result list
\end{myverbatim}

\textbf{Anti-Tribute Notification.} When the conditions for anti-tribute are satisfied, the server notifies all players of the number of players performing anti-tribute (\texttt{antiNums}) and their corresponding seat indices (\texttt{antiPos}).
\begin{myverbatim}
'type': 'notify',
'stage': 'anti-tribute',
'antiNums': 2,	# Number of players performing anti-tribute
'antiPos': [0, 2]	# List of seat numbers of anti-tribute players
\end{myverbatim}

\textbf{Back-Tribute Notification.} After the back-tribute phase is completed, the server broadcasts the back-tribute results to all players. The field \texttt{result} specifies the exchanged cards between tribute-paying and back-tribute players.
\begin{myverbatim}
'type': 'notify',
'stage': 'back',
'result': [[3, 0, 'S2']]	# List of back-tribute results
\end{myverbatim}

\textbf{Episode Over.} After each round terminates, the server broadcasts the round summary to all players. This includes the order in which players emptied their hands (\texttt{order}), the level of the current round (\texttt{curRank}), and the remaining cards held by unfinished players (\texttt{restCards}). The corresponding message format is as follows:
\begin{myverbatim}
'type': 'notify',
'stage': 'episodeOver',
'order': [0, 1, 2, 3],	# Finishing order
'curRank': 'A',	# Current level
'restCards': [[3, ['C2']]]	# List of remaining cards
\end{myverbatim}

\textbf{Game Over.} After the game session ends, all players receive the game-level summary, including the current number of completed games (\texttt{curTimes}) and the predefined maximum number of games (\texttt{settingTimes}). The message format is given below:
\begin{myverbatim}
'type': 'notify',
'stage': 'gameOver',
'curTimes': 1,	# Current number of completed games
'settingTimes': 1	# Configured number of games
\end{myverbatim}

\textbf{Game Result.} Upon game termination, the server broadcasts the final game outcome to all players. Specifically, \texttt{victory} denotes the winning team, and \texttt{victoryRank} records the final levels of both teams. The corresponding message format is shown as follows:
\begin{myverbatim}
'type': 'notify',
'stage': 'gameResult',
'victory': 0,	# Winning team
'victoryRank': ['A', 'K']	# Final levels of both teams
\end{myverbatim}

Below, we present the messages targeted at the specific player who is required to execute an action.

\textbf{Play Request.} For the acting player, the server provides comprehensive information, including the player's private hand cards, the remaining card counts of all players, the levels of both teams, the level of the current round, the seat index of the player holding the highest action, the current highest action, and the set of available actions. Based on these information, researchers can construct diverse observation representations for learning-based agents. The message format is as follows:
\begin{myverbatim}
'type': 'act',
'handCards': ['S2', 'H2', ...],		# Current player's hand cards
'publicInfo': [		# Public information
{'rest': 22},	# Rest indicates remaining card count
{'rest': 23},
{'rest': 23},
{'rest': 27}
],
'selfRank': 'K',	# Own team's level
'oppoRank': '9',	# Opponent team's level
'curRank': 'K',		# Current level
'stage': 'play',
'curPos': 2,		# Current player's seat number
'curAction': ['Bomb', 'A', ['HA', 'HA', 'CA', 'DA']],	# Current player's action
'greaterAction': ['Bomb', 'A', ['HA', 'HA', 'CA', 'DA']],	# Highest action
'greaterPos': 2,	# Seat number of player with highest action
'actionList': [		# Available action list
['PASS', 'PASS', 'PASS'],
['Bomb', '9', ['H9', 'H9', 'C9', 'D9']],
...
],
'indexRange': 21	# Maximum index value
\end{myverbatim}

\textbf{Tribute Request.} During the tribute phase, the server provides the acting player with its private hand cards, the levels of both teams, the level of the current round, and the set of available tribute actions. The message format is given below:
\begin{myverbatim}
'type': 'act',
'handCards': ['H3', 'D3', ...],		# Current player's hand cards
'selfRank': '2',	# Own team's level
'oppoRank': '9',	# Opponent team's level
'curRank': '9',		# Current level
'stage': 'tribute',
'actionList': [['tribute', 'tribute', ['D2']]],	# Available tribute action list
'indexRange': 0		# Maximum index value
\end{myverbatim}

\textbf{Back-Tribute Request.} After receiving a tribute card, the recipient player is required to perform a back-tribute action. In this phase, the server provides the player with its private hand cards, the levels of both teams, the level of the current round, the seat index of the tribute-paying player, the tribute card, and the set of available back-tribute actions. The corresponding message format is shown below:
\begin{myverbatim}
'type': 'act',
'handCards': ['H2', 'S3', ...],		# Current player's hand cards
'selfRank': '5',	# Own team's level
'oppoRank': '9',	# Opponent team's level
'curRank': '9',		# Current level
'stage': 'back',
'tributePos': 3,	# Seat number of player who paid tribute
'tribute': 'S2',	# The card offered as tribute
'actionList': [		# Available back-tribute action list
['back', 'back', ['H2']],
['back', 'back', ['S3']],
...
],
'indexRange': 11	# Maximum index value
\end{myverbatim}

\section{API}
\label{sec:api_design}
The GuanDan simulator provides independent action-upload API for each player, which makes it compatible with both humans and AI agents. Below, we present example code snippets illustrating how actions are submitted during the card play, tribute, and back-tribute phases, respectively.

\begin{myverbatim}
type: 'PLAY',
data: {
	roomID: self.room_id,
	player: self.player_idx,
	act: msg['actionList'][act_index]}
\end{myverbatim} 

\begin{myverbatim}
type: 'TRIBUTE',
data: {
	roomID: self.room_id,
	player: self.player_idx,
	act: msg['actionList'][act_index]}
\end{myverbatim} 

\begin{myverbatim}
type: 'PAYTRIBUTE',
data: {
	roomID: self.room_id,
	player: self.player_idx,
	tributePos: msg['tributePos']
	tribute: msg['tribute']
	act: msg['actionList'][act_index]}
\end{myverbatim} 

In the above code snippets, \texttt{roomID} and \texttt{player} respectively refer to the indices of the current game instance and the acting player. The input \texttt{act\_index} specifies the index of the selected action within the legal action set, which may be produced by either human decision-making or AI models. The variable \texttt{msg} represents the information broadcast by the simulator to each player at the current time step. This API design facilitates efficient human-AI evaluation, and supports seamless integration of diverse GuanDan AI agents developed using game theory, RL, and LLM, while also enabling the self-play training paradigm.

\section{Inference Time Evaluation}
\label{sec:inference_time_evaluation}
In addition to the pair-wise evaluations among all GuanDan AI agents and the human-AI evaluations between human players and all learning-based AI agents, we report the computational time cost of each AI agent during action selection. As shown in Tab.~\ref{tab:inference_time_evaluation}, Rule three and Rule Four achieve 926 and 1005 time steps per second, respectively, indicating extremely low computational overhead. This efficiency stems from their simple handcrafted rule-based designs that cover only limited decision-making scenarios in GuanDan, which consequently leads to inferior performance compared to other agents. 

\begin{table*}[h]
	\caption{The computational time cost of all GuanDan AI agents during action selection.}
	\label{tab:inference_time_evaluation}
	\centering
	\resizebox{0.8\linewidth}{!}{
		\begin{tabular}{c|c|c|c|c|c|c|c}
			\toprule
			AI Agent & GS2 & SDMC & DanZero & Rule One & Rule Two & Rule Three & Rule Four \\
			\midrule
			Time (time steps/s) & 102 & 136 & 384 & 56 & 30 & 926 & 1005 \\
			\bottomrule
		\end{tabular}
	}
\end{table*}

Although Rule One and Rule Two exhibit lower inference throughput (only 56 and 30 time steps per second, respectively), their weak performance against learning-based agents demonstrates that data-driven AI techniques are more promising to address large-scale imperfect information games. For learning-based agents, SDMC and DanZero are both developed based on the DMC method and the self-play training paradigm. However, SDMC incorporates soft action sampling to mitigate exploitability during action selection, which results in slower inference compared to DanZero. Moreover, GS2 extends SDMC with real-time search, leading to additional computational overhead and further reducing inference speed. These results highlight the need for more efficient real-time search mechanisms to improve inference efficiency in future research.

\end{document}